\documentclass[twoside]{article}

\usepackage[accepted]{aistats2023}
\usepackage[round]{natbib}
\usepackage{color}
\renewcommand\refname{}
\usepackage{amssymb}

\usepackage{graphicx}
\usepackage[ruled,vlined,linesnumbered,]{algorithm2e}
\usepackage{float}
\begin{document}

%

%

\twocolumn[

\aistatstitle{Jump-Diffusion Langevin Dynamics for Multimodal Posterior Sampling}

\aistatsauthor{ Jacopo Guidolin \And Vyacheslav Kungurtsev \And  Ondřej Kuželka }

\aistatsaddress{ University of Padua\textsuperscript{1} \And Czech Technical University\textsuperscript{2} \And Czech Technical University\textsuperscript{2} } ]

\begin{abstract}
  Bayesian methods of sampling from a posterior distribution are becoming increasingly popular due to their ability to precisely display the uncertainty of a model fit. Classical methods based on iterative random sampling and posterior evaluation such as Metropolis-Hastings are known to have desirable long run mixing properties, however are slow to converge. Gradient based methods, such as Langevin Dynamics (and its stochastic gradient counterpart) exhibit favorable dimension-dependence and fast mixing times for log-concave, and ``close'' to log-concave distributions, however also have long escape times from local minimizers. Many contemporary applications such as Bayesian Neural Networks are both high-dimensional and highly multimodal. In this paper we investigate the performance of a hybrid Metropolis and Langevin sampling method akin to Jump Diffusion on a range of synthetic and real data, indicating that careful calibration of mixing sampling jumps with gradient based chains significantly outperforms both pure gradient-based or sampling based schemes.
\end{abstract}

\section{INTRODUCTION}
\footnotetext{The author acknowledges the support of the ``Erasmus+'' Programme.}
\footnotetext[2]{The authors acknowledge the support of the OP VVV funded project 
CZ.02.1.01/0.0/0.0/16\_019/0000765 ``Research Center for Informatics''. The access to the computational infrastructure of the OP VVV funded project CZ.02.1.01/0.0/0.0/16\_019/0000765 ``Research Center for Informatics'' is also gratefully acknowledged.}

One of the key tasks in machine learning is trying to estimate a complex and unknown model with an approximated surrogate. Most of the time the latter one will be a parametric model i.e. a model heavily dependent on a finite number of learnable parameters.

In this work we will concentrate on the setting of Bayesian inference. In particular our interest is to investigate effective techniques to productively explore the posterior distribution on the space of the parameters of any given parametric model; \citep{MacKay95}, \citep{nealbook}.

There are two main paradigms that are commonly followed in order to perform this task. The first one being associated with techniques from the wide branch of optimization and the second being sampling, specifically sampling by means of Markov Chain Monte Carlo based methods \citep{barbu2020monte}.

Broadly speaking, sampling algorithms, which use a known proposal distribution and careful probabilistic sample acceptance-rejection criteria, are known to reliably mix to a wide range of posteriors. However, the mixing rate, or convergence in terms of probability measures, is slow, and scales poorly with the parameter dimension; \citep{metroconvtweedie}, \citep{metroconv2}. 

By contrast, optimization based, or gradient-informed methods, exhibit fast ergodicity for log-concave potentials (see, e.g. \citet{ref1}, \citet{ref2} and \citet{ref3}). In addition to that, while they are long-run ergodic, practically they may be seen to exhibit ``mode seeking'' behavior, wherein one mode in a multi-modal distribution is sampled for a large portion of the chain, leaving the rest of the parameter space less informed. This was more comprehensively studied by \citet{samplopt}, who showed that for a specific class of nonconvex potentials, the dimensional dependence of sampling is superior to that of gradient-based methods. 
Recently, however, there has been interest in large scale multimodal posterior models.
In particular performing Bayesian inference on the parameters of a Neural Network can provide desirable properties. Bayesian Neural Networks (BNNs) can bring, along with reliable uncertainty estimates and an intrinsic protection against overfitting, natural support for active learning, continual learning, and decision-making under uncertainty \citep{bayespost}.
Yet, the posteriors characterizing BNNs exhibit both significant multi-modality, which should benefit sampling-based methods, while at the same time having a large parameter dimension (and hence exhibit favorable performance for gradient based samplers), and the modes have standalone significance for inference (i.e., for MAP inference). 

In this paper, we introduce a procedure based on performing Jump Diffusions on a singular posterior, rather than explicitly jumping across components of a mixture. The intention is to achieve the advantages of point-based and gradient-based sampling simultaneously: periodic jumps ensure sufficient exploration of the parameter space and subchains of Langevin inform the precise details of the, generally, most significant modes. We shall demonstrate that indeed, on both synthetic and real data sets, the combination of the two procedures performs better than either point-based or gradient-based sampling procedures alone.

\section{BACKGROUND}
In this Section we introduce two standard representatives of the two paradigms of sampling as well as introduce our jump diffusion based procedure.

\subsection{Metropolis Hastings Algorithm}

Metropolis-Hastings (MH) is a reliable algorithm especially in the context of unnormalized probabilities. It allows us to draw samples from a desired probability distribution $P$ of density $p(x)$ if we have access to a function $f(x)$ such that $f \propto p$, \citep{barbu2020monte}.

The main idea of the algorithm is to take any method that iteratively jumps between different states and adding to the latter an additional \emph{acceptance step} between the jumps.
The purpose of the acceptance step is to ``correct" the initial algorithm in such a way that the Markov Chain generated by the updated algorithm will satisfy the detailed balance equation. In particular, given that $\mathcal{K}$ is the transition probability of the chain, that is:
\begin{equation*}
    \mathcal{K}\left( X,Y \right) p(X) =  \mathcal{K}\left( Y,X \right) p(Y).
\end{equation*}
The standard way of proceeding is to start with an inexpensive proposal distribution that allows us to conditionally sample from it rather inexpensively, only then correct this sampling by applying MH. Each of these proposed samples can then be either accepted or rejected according to a particularly designed \emph{acceptance probability}.


It has to be noted that the more we move into high dimensional settings, and the more the geometry of the posterior distribution gets complex, the more an accurate design for the proposal distribution plays a key role on yielding an effective rate of convergence for MH. Formally, the mixing rate of MH depends on the worst-case ratio of the desired and proposal distributions. This factor typically scales unfavorably with the dimension; \citep{metroconvtweedie}, \citep{metroconv2}.

We can see pseudocode for the Metropolis-Hastings with Independence Metropolis Sampler design in Algorithm \ref{alg:metr}, \citep{barbu2020monte}. 

\begin{algorithm}
\KwIn{target distribution $p(\cdot)$, current state of the chain $\theta_t$, a proposal probability $q(\cdot)$}
\KwOut{the next state of the chain $\theta_{t+1}$}
\SetAlgorithmName{Metropolis-Hastings Algorithm,}{}{}

    Generate a new proposal state $\Bar{\theta}$ by sampling from $q$
    
    Calculate the acceptance probability:
    \begin{equation*}
        p_{a}(\theta, \Bar{\theta}) = \min \left( 1, \frac{q(\theta)}{q(\Bar{\theta})} \frac{p(\Bar{\theta})}{p(\theta)}\right)
    \end{equation*}

    Accept the new state with probability $p_{a}(\theta,\Bar{\theta})$ and set $\theta_{t+1} = \Bar{\theta}$, otherwise $\theta_{t+1} = \theta_t$.
\caption{}
\label{alg:metr}
\end{algorithm}

\subsection{Stochastic Gradient Langevin Dynamics}

Let $\mathcal{M}$ be a parametric model with $\theta$ as vector of parameters, $\theta \in \Theta$. Let $p\left(\cdot\right)$ be a prior distribution on $\Theta$. Let $\mathcal{X}$ be the input space of our model and, given $x \in \mathcal{X}$, let $p\left(x \mid \theta \right)$ be the density of the quantity $x$ if our model has $\theta$ as parameter vector. Given a set of $n$ data items the update rule for the parameters is the following:


\begin{equation*}
\begin{aligned}
\Delta \theta_t &= \frac{\varepsilon_t}{2} \Bigg( \underbrace{\nabla \log\left(p\left(\theta_t\right)\right)}_{\substack{\text{priors gradient}}} \\
                & \quad + \frac{N}{n} \sum_{i=1}^n \underbrace{\nabla \log\left(p( x_i \mid \theta_t ) \right)}_{\substack{\text{likelihood gradients}}} \Bigg) \\
                & \quad + \underbrace{\nu_t}_{\substack{\text{noise}}}
\end{aligned}
\end{equation*}
where $\nu_t \sim N\left(0, \varepsilon_t\right)$.

It is important to note that the gradient step sizes and the standard deviation of the injected noise are balanced in such a way that the variance of the noise matches the variance of the posterior.

Naturally, discretization of Langevin dynamics, as necessary for practical computation, introduces bias. Performing a Metropolis-Hastings acceptance-rejection step, however, corrects this. 

One other possible approach is to use a scheduled stepsize decay, as seen in \citet{wnt}, which is also associated with a “stochastic gradient” approach of using subsamples. This approach slows the per sample convergence rate, however, and appears to be additionally mode-seeking.

Langevin dynamics has been studied extensively in regards to its mixing properties. The associated SDE model appears in a number of physical phenomena as well, and thus the Langevin type equation has received considerable analysis~\cite{stroock1997multidimensional}. As an algorithm for sampling, its interest has increased due to the improved dimensional dependence of the mixing rate. This is especially the case of log-concave potentials, for which the performance is akin to gradient descent for convex functions, convergence is fast. At the same time, however, the gains in local accuracy from using the gradient are contrasted with the local inertia biasing against global exploration of the parameter space. Namely, the wells corresponding to attraction basins for the Langevin dynamics also exhibit long escape times, and thus slow tendency to explore outside of a contemporaneous mode, \citep{chen2020stationary}.

\subsection{Jump-Diffusion}

Our method is inspired by the concept of Jump-Diffusions, also known as Hybrid Switching Diffusions in the literature \citep{bookhybrid}. Jump-Diffusions are a particular class of diffusion processes involving multiple diffusions, or SDEs with a drift and a Wiener process, together with a Poisson jump process. The jump process randomly switches between the different diffusions. 

Aside from describing some complex stochastic systems in physics, jump diffusions have been applied to sample from mixtures. They were analyzed for its statistical potential in~\citep{green1995reversible}. Additional case studies appear in~\citep{grenander1994representations}. Broadly speaking, the procedures have potential for successfully sampling from posterior distributions with mixtures of notably different distributions, with some notable success in applications such as image segmentation~\citep{han2004range}. Quality performance, however, typically requires well informed data-driven priors. 

The use of jump-diffusions for single potentials is, to the best of the authors' knowledge, unexplored, despite its natural intuitive potential. Below we formally introduce our particular procedure inspired by the hybrid jump diffusion literature.


\section{JUMP-DIFFUSION LANGEVIN DYNAMICS}

We now propose our jump-diffusion algorithm for single posterior sampling. The method combines MH with adjusted Langevin. In particular, for some number of samples, it performs a chain of adjusted Langevin. At the same time, after a number of such samples, the number being itself sampled from a Poisson distribution, an MH step occurs: a proposal distribution is sampled and then accepted or rejected. 

We define the constant stepsize of the Langevin sequence by $\varepsilon$ and note that the frequency of jumps is regulated by a Poisson process, $\tau \sim \text{Pois}(\lambda)$, where $\lambda$ can be considered a tunable hyperparameter.

The underlying intuition is to prevent the Langevin diffusion from getting stuck on certain modes by allowing the process to jump to a different state without making use of first order information. We intend for the chain to, as such, have high resolution for the most significant modes, while at the same time ensuring that all nodes are explored, and the broad uncertainty as well (but at coarser resolution). We shall see that this is precisely the behavior we see in practice.

One can consider running parallel, or multiple sequential, chains with different starting points. There are significant problems with this approach, however. First, the lack of guarantees, since the jump process is not Metropolis-regulated, as in our case. This would introduce bias. Second, this would depend on the choice of the associated hyperparameters -- there should be a nice correspondence between the number of modes and the number of starting points, otherwise we can't expect such a method to perform well. By contrast, with a large enough sample size, the procedure we introduce should effectively characterize the posterior, with minimal hyperparameter dependence.

\begin{algorithm}[t]
\KwIn{target distribution $p(\cdot)$, current iteration $t$, current state of the chain $\theta_t$, a proposal probability $q(\cdot)$, next jumping iteration $\Tilde{t}$, jump rate $\lambda$}
\KwOut{The next state of the chain $\theta_{t+1}$, the next jumping iteration $\Tilde{t}$}
\SetAlgorithmName{Our Algorithm,}{}{}

  \uIf{$t = \Tilde{t}$}{
    Generate a new proposal state $\Bar{\theta}$ by sampling from $q$
    
    Calculate the acceptance probability:
    \begin{equation*}
        p_{a}(\theta_t, \Bar{\theta}) = \min \left( 1, \frac{q(\theta_t)}{q(\Bar{\theta})} \frac{p(\Bar{\theta})}{p(\theta_t)}\right)
    \end{equation*}

    Accept the new state with probability $p_{a}(\theta_t,\Bar{\theta})$ and set $\theta_{t+1} = \Bar{\theta}$, otherwise $\theta_{t+1} = \theta_t$.
    
    Sample $i \sim \text{Pois}(\lambda)$
    
    $\Tilde{t} = \Tilde{t} + i$}
  \uElse{
  
  Calculate the Langevin step update:
    \begin{equation*}
    \begin{aligned}
         \Delta \theta_t &= \frac{\varepsilon}{2} \Bigg( \nabla \log(p(\theta_t)) \\
         &\quad+ \sum_{i=1}^N \nabla \log(p( x_i \mid \theta_t))\Bigg) \\
         &\quad+ \nu_t
    \end{aligned}
    \end{equation*}
    
    Set $\theta^*=\theta_t + \Delta\theta_t$.
    
    Calculate the acceptance probability:
    \begin{equation*}
        p_{a}(\theta_t, \theta^*) = \min \left( 1, \frac{q(\theta_t)}{q(\theta^*)} \frac{p(\theta^*)}{p(\theta_t)}\right)
    \end{equation*}

    Accept the new state with probability $p_{a} ( \theta_t, \theta^* )$ and set $\theta_{t+1} = \theta^*$, otherwise $\theta_{t+1} = \theta_t$.}
\caption{}
\end{algorithm}


The use of MH-regulated jumps is facilitated by a good choice of the proposal probability. This can be done in a data-driven way, with some form of pretraining. Or, alternatively, with sufficiently large variance, standard Gaussians should work sufficiently well~\citep{bayespost}.

\subsection{Theoretical Properties}
The seminal reference on jump diffusions is the monograph~\cite{bookhybrid}. The canonical system is a set of stochastic diffusions and a Poisson jump process that switches between them. Formally,
\[
dX(t) = b(X(t),\alpha(t)) dt + \sigma(X(t),\alpha(t))dW 
\]
where $\alpha(t)\in \mathcal{N}\subset\mathbb{N}$ is a set of integers identifying distinct diffusion processes with (potentially state dependent) transition matrix $q_{ij}(x)$.

Theorem 4.4~\cite{bookhybrid} establishes weak convergence of the probability density corresponding to the stochastic process to a stationary distribution corresponding to a mixture of the stationary distributions for the set of diffusions. Since in our case, the diffusions are the same, this corresponds to the desired stationary distribution. 

The use of Metropolis adjusted Langevin for the chains results in an unbiased simulation of the diffusion, and as such the discretization maintains the ergodic properties of the SDE. Exponential convergence under the log-concave condition for the potentials is established in~\cite{roberts1996exponential}. Thus, during the procedure, any samples taken within a basin with a locally strongly convex potential are converging geometrically in probability to the distribution associated with that mode. Thus mixing is particularly fast in the regions corresponding to potential modes, which is broadly desired as accurate characterization of posterior modes is more significant for the user than the rest of the phase space.

Taken together, we can conglomerate the conclusion of the theoretical results as follows: 
\begin{itemize}
    \item As Metropolis-adjusted discretized Langevin maintains the ergodicity of the Langevin sampling procedure, and jump diffusions are known to be ergodic, we expect the procedure to converge asymptotically, in standard probability distances, to the desired posterior
    \item When the chain is exploring a mode that is locally strongly log-concave, the localized parameter region corresponding basin of attraction for that mode in terms of gradient flows exhibits geometric ergodicity, i.e., exponential convergence towards the mode.
\end{itemize}
With these in mind, we can see how depending on the landscape, we should see how broadly this procedure should result in faster mixing than either Langevin or Metropolis MCMC alone.

\section{EXPERIMENTS}

Here we present the findings of our numerical experiments comparing the sampling methods described above. In standard practice, we shall present results with both low-dimensional models whose posteriors we can display visually as well as more coarse comparisons for large-dimensional models.

\subsection{Low-Dimensional Problems}

\subsubsection{One Dimensional Mixture}
We begin with some standard one-dimensional examples. In particular, we clearly show that if the distribution we aim to estimate is multimodal, it can be difficult to estimate by means of pure Langevin Dynamics even in the one-dimensional case.

\begin{figure}[tb!]
\centering
    \includegraphics[scale=0.35]{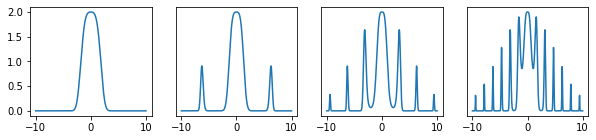}
    \\[\smallskipamount]
    \hspace{-7.6pt}\includegraphics[scale=0.35]{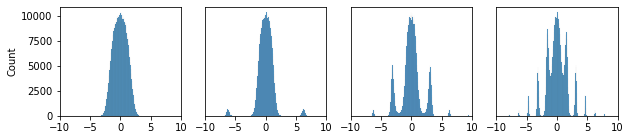}
    \\[\smallskipamount]
    \hspace{-7.6pt}\includegraphics[scale=0.35]{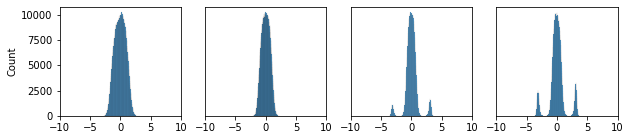}
    \\[\smallskipamount]
    \hspace{-7.6pt}\includegraphics[scale=0.35]{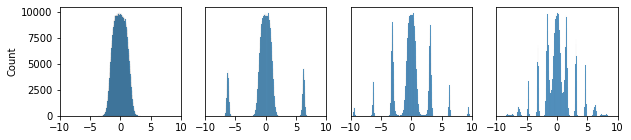}
\caption{Target distribution and its estimations. Target, Metropolis-Hastings, Langevin and Ours on the rows; $\alpha \in \{-2,-1,0,1\}$ on the columns, respectively. Every method was run for 1M iterations.}
\label{fig:dunes}
\end{figure}

In particular we assumed the posterior distribution to be of form $p(x) = \frac{1}{Z} e^{-E(x)}$ where
\begin{equation}
    E(x) =x^2(\sin^2({2^{\alpha}x)} + \delta)
\end{equation}
with $\delta = 2\cdot10^{-2}$ and $\alpha \in \{-2,-1,0,1\}$.
We tried to estimate this hand-crafted distribution by means of Metropolis-Hastings, Langevin Dynamics and Our method. All methods were run for 1M iterations reaching the results shown in Figure \ref{fig:dunes}.
As we can see, the multimodality of the distribution is barely captured when using pure Langevin Dynamics.

\subsubsection{Two Dimensional Mixture}
Next we generated a two dimensional dataset in such a way that the resulting posterior distribution was multimodal with distinct and slightly separated modes. We suspect that these traits will prompt Stochastic Gradient Langevin Dynamics to produce incorrectly estimated posterior distributions.

\begin{figure}[tb!]
  \centering
  \includegraphics[scale=0.35, trim = 2cm 0 0 0.1cm]{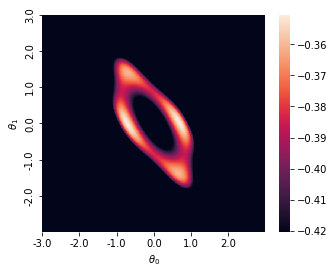}
  \includegraphics[scale=0.35, trim = 0.4cm 0 0 0]{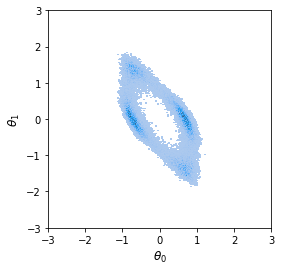}
  \caption{True posterior distribution (left) and posterior distribution estimated using Metropolis-Hastings (right).}
  \label{fig:truepot2}
\end{figure}

We let two parameters, $\theta_0$ and $\theta_1$, were sampled from a Gaussian distribution: $\theta_0,\theta_1 \sim N\left(0, 1 \right)$. Then we sampled 100 data points $\left\{ x_1, ... , x_{100} \right\}$ with:
\begin{equation*}
    x_i \sim \frac{1}{2} N \left( \theta_0^2, 1 \right) + \frac{1}{2} N \left( \left(\theta_0 + \theta_1\right)^2, 1 \right)
\end{equation*}
where $\theta_0 = 0$, $\theta_1 = 1$.

We know the true posterior distribution in the parameter space and so we are able to plot it. In particular, working with log-probabilities:
\begin{equation*}
\begin{aligned}
 L\left(\theta_0, \theta_1\right) &= \sum_{i = 1}^{100} \log \left( \frac{1}{2} e^{\frac{\left( x_i - \theta_0^2\right)^2}{2}} + \frac{1}{2} e^{\frac{\left( x_i - (\theta_0+\theta_1)^2\right)^2}{2}} \right)\\
                                    & \quad+ \left(\frac{-\theta_0^2}{2}\right) + \left(\frac{-\theta_1^2}{2} \right)
\end{aligned}
\end{equation*}
resulting in the plot seen in Figure \ref{fig:truepot2}.

In addition, in Figure \ref{fig:truepot2}, we show the empirical distribution generated by MH. Even if 1M iteration can be said to be low, the resulting posterior distribution is sound, the multimodality of the posterior is correctly captured and the ratio between the modes seems to be preserved.


As described in \citet{wnt}, and due to the similar assumptions, we ran the Stochastic Gradient Langevin process with a batch size of 1, for 10000 sweeps through the whole dataset (for a total of 1M iterations) and with decaying learning rate from $10^{-2}$ to $10^{-4}$ regulated by $\varepsilon_t = a\left(b+t\right)^{\gamma}$ where $a=0.2$, $b = 231$ and $\gamma = 0.55$.


\begin{figure}
\centering
    \hspace{-15pt}\includegraphics[width=.19\textwidth]{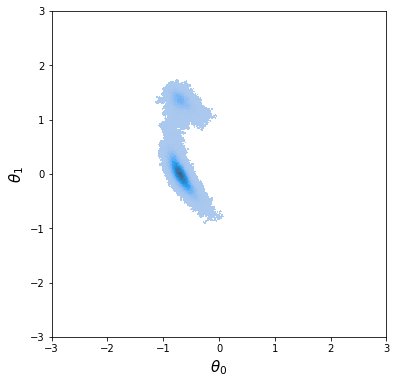}
    \includegraphics[width=.19\textwidth]{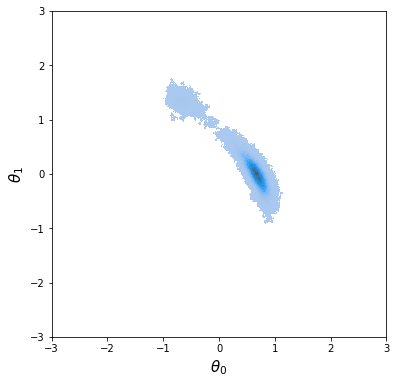}
    \\[\smallskipamount]
    \hspace{-15pt}\includegraphics[width=.19\textwidth]{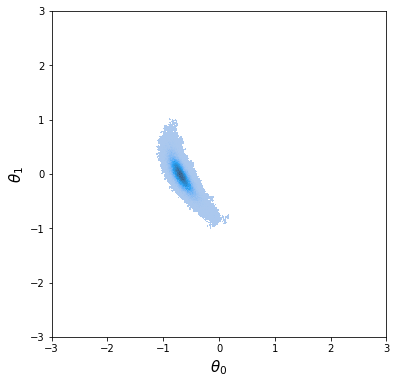}
    \includegraphics[width=.19\textwidth]{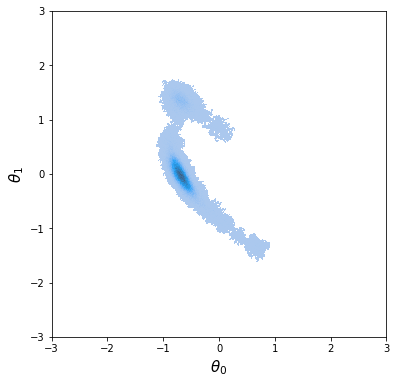}

\caption{Examples of unsuccessfully estimated distributions using Stochastic Gradient Langevin Dynamics.}
\label{fig:lang_group}
\end{figure}

By examining the posterior distribution estimated by SGLD (Figure \ref{fig:lang_group}), we see the similar mode-seeking behavior. Due to the further separation of the modes and the higher complexity of the posterior, the resulting estimated distribution has notably truncated support. The method did not manage to visit the parameter space extensively and the obtained distribution can be labeled as unsatisfactory.

\begin{figure}[tb!]
\centering
    \includegraphics[width=.11\textwidth]{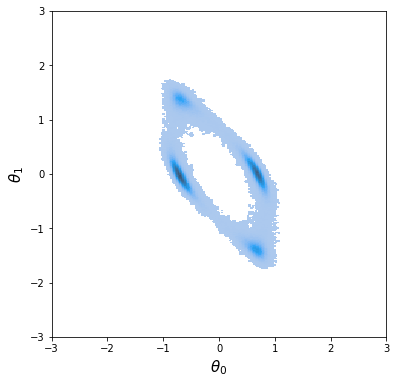}
    \includegraphics[width=.11\textwidth]{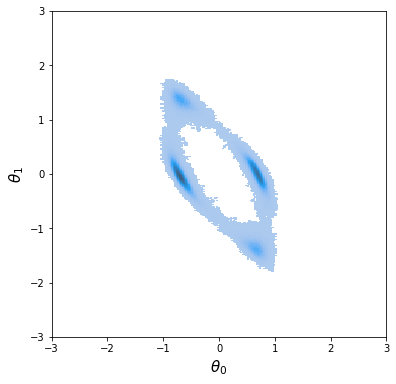}
    \includegraphics[width=.11\textwidth]{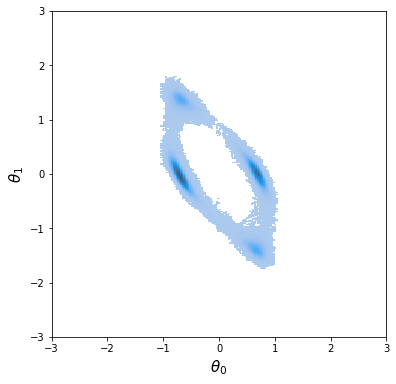}
    \includegraphics[width=.11\textwidth]{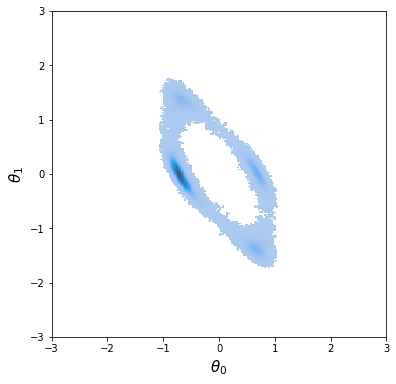}
    \\[\smallskipamount]
    \includegraphics[width=.11\textwidth]{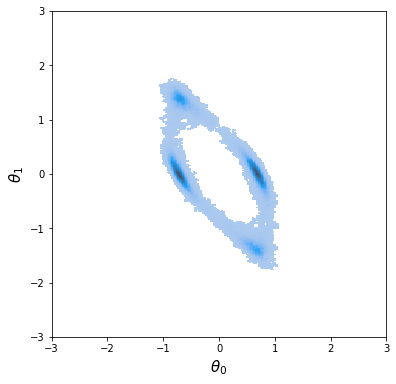}
    \includegraphics[width=.11\textwidth]{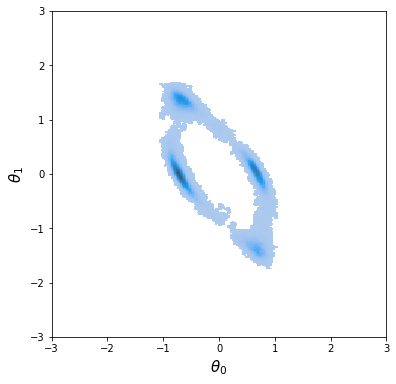}
    \includegraphics[width=.11\textwidth]{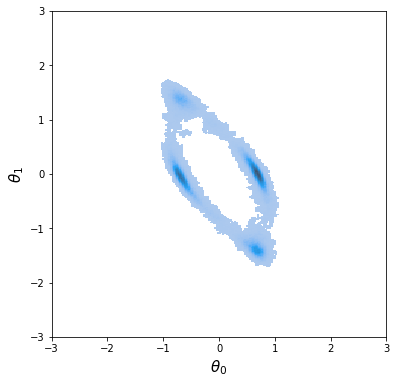}
    \includegraphics[width=.11\textwidth]{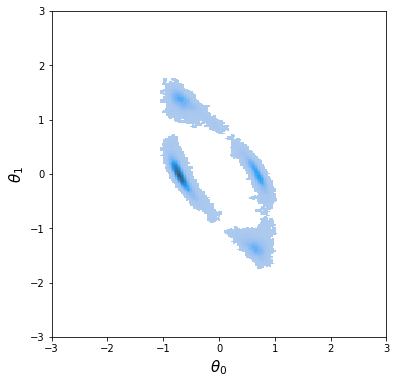}
    \\[\smallskipamount]
    \includegraphics[width=.11\textwidth]{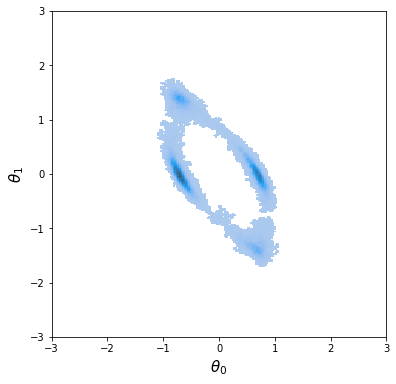}
    \includegraphics[width=.11\textwidth]{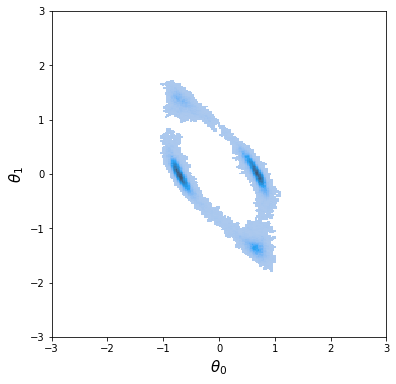}
    \includegraphics[width=.11\textwidth]{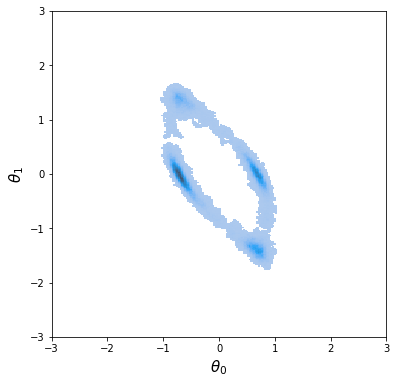}
    \includegraphics[width=.11\textwidth]{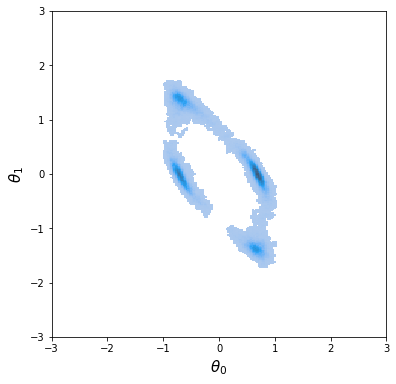}
\caption{Distributions estimated with Our method with $\lambda \in \{50, 100, 200, 1000\}$ on the columns and $\varepsilon \in \{5 \cdot 10^{-5} , 1 \cdot 10^{-5}, 5 \cdot 10^{-6}\}$ on the rows, respectively.}
\label{fig:jlang_lambdas}
\end{figure}

Next, in considering our Jump Langevin procedure, we extensively tested multiple combinations of hyperparameter settings, involving varying stepsizes and jump frequency $\lambda$, some of which are shown in Figure \ref{fig:jlang_lambdas}.
We can say that our method seems to be insensitive to the choice of the jump $\lambda$ parameter. On the stepsize does affect the results, but such is standard in using Langevin discretizations. 

Overall we can say that almost all the combinations of settings used in Figure \ref{fig:jlang_lambdas} yielded satisfactory estimation of the posterior, managing to correctly express its multimodality while producing a precise characterization of the significant modes.

\subsubsection{Synthetic Highly-Multimodal Posterior}

Finally, among low dimensional examples, we now produce an even more multimodal distribution to test the sampling algorithms under even more challenging conditions.

For these experiments, we did not generate a dataset, instead we assumed direct access to first and zeroth order information of the posterior distribution.

\begin{figure}[tb!]
  \centering
    \hspace*{-10pt}\includegraphics[scale = 0.4,  trim = 0.2cm 0 0 0.2cm]{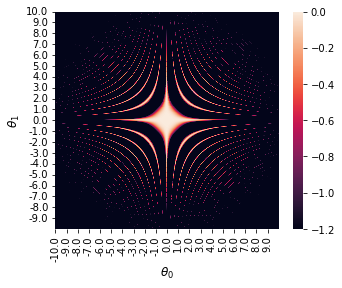}
    \includegraphics[scale = 0.4, trim = 0.2cm -0.6cm 0 0]{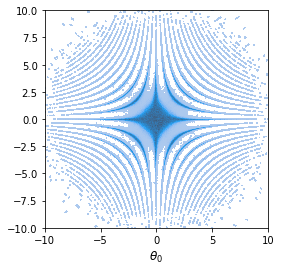}
  \caption{True posterior distribution (left) and posterior distribution estimated with Our Method (right).}
  \label{fig:truenjlang}
\end{figure}

In particular we assumed the posterior distribution to be of form $p(x) = \frac{1}{Z} e^{-E(x)}$ where
\begin{equation}
    E(x) = (x^2 + y^2)(\sin^2{xy} + \delta)
\end{equation}
with $\delta = 10^{-2}$.
The obtained synthetic posterior proves to be highly multimodal, with dense and separated modes as can be seen in Figure \ref{fig:truenjlang}.

For a sound estimation of the posterior we generated a chain with 50M+ states using Metropolis-Hastings and we removed the initial “burn-in” chunk, obtaining a chain with 50M states.

Trying to properly estimate this distribution using pure Langevin Dynamics was unachievable, the resulting distributions were almost exclusively unimodal and heavily dependent on the starting point.

On the other hand, when trying to estimate the distribution with Our method, we obtain the result showing in Figure \ref{fig:truenjlang}. The presented distribution was obtained after keeping the last 50M states of a chain with 50M+ states generated with our method.

\begin{figure}[tb!]
\centering
    \includegraphics[width=.24\textwidth]{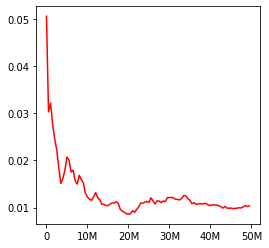}
    \includegraphics[width=.24\textwidth]{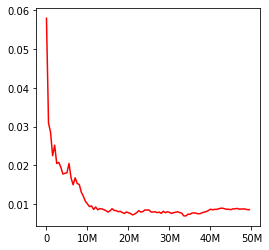}

\caption{Plot of Wasserstein Distance for $\theta_0$ and $\theta_1$ distributions, respectively.}
\label{fig:wd_group}
\end{figure}

We also measure the “difference", by means of the (W1) Wasserstein Distance,  between the distribution generated by the full MH chain and the distribution generated with our method, as the number of chain-states grew. We show the satisfactory results we obtain in Figure \ref{fig:wd_group}.

\subsection{High-dimensional setting}

For the experiments in a high dimensional setting we tested our method on Bayesian Neural Networks.

In particular, we take into consideration a binary classification problem. The dataset we used is the \emph{fourclass} dataset found in the LIBSVM library.
For this task we chose to use a fully connected neural network with 3 hidden layers of 8, 8 and 4 neurons, respectively. The output layer had only 1 neuron with a sigmoid as activation function.

We pretrained our network with SGD in order to gain some previous knowledge about the parameters of the network, specifically we center the priors on the optimal parameters, but use a relatively high standard deviation, in the context of Neural Network parameters, of 0.5.

We run three separate chains using Metropolis-Hastings, Langevin Dynamics and Our method for $3 \cdot 10^6$ iterations, with a stepsize of $3 \cdot 10^6$ for both gradient based methods and $\lambda=500$ for Our method.

\begin{figure}[tb!]
\centering
    \includegraphics[scale=0.575]{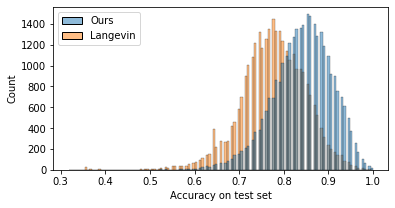}
\caption{Distribution of accuracies for the sampled chain states. “Holes” on the x axis are caused by the finite number of elements in the test set.}
\label{fig:networkcomp}
\end{figure}

To test the three methods we we took one sample 30k times from each chain. For each sample of the parameter vector, we performed inference on the test set and evaluated the performance by computing the accuracy of the predicted and actual label values.
From our findings, the Metropolis-Hasting results were both noisy (i.e., exhibiting large variability across multiple runs) and always significantly underperformed the other two methods.  We can see a comparison between pure Langevin Dynamics and Our method in Figure \ref{fig:networkcomp}.

\subsection{Additional Experiments}

We want to provide a systematic test that highlights the comparison between Langevin Dynamics and Jump Langevin (our method) on some increasingly multimodal posteriors. To do that we synthetically generated four unidimensional posterior distributions with, respectively, 2, 8, 16 and 32 equal and separated modes.

We then created a baseline in order to compare our methods. To do that, for every one of the four posteriors, we generated a chain with 10M states using the Metropolis-Hastings algorithm. We then generated, for each different posterior, one chain with 10M states using Langevin Dynamics and one chain with 10M states using Our method. To conclude we observed the evolution of the measured Wasserstein Distance as the number of chain states increased.

\begin{figure*}[tb!]
\centering
    \includegraphics[scale=0.36]{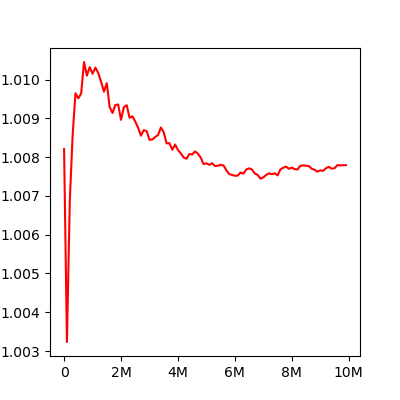}
    \includegraphics[scale=0.36]{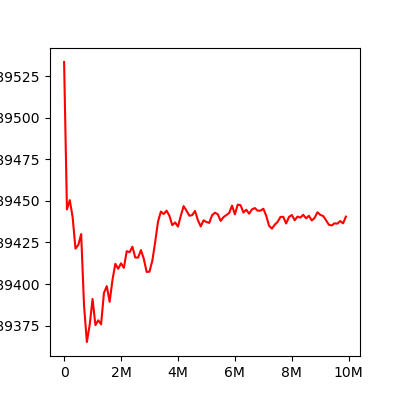}
    \includegraphics[scale=0.36]{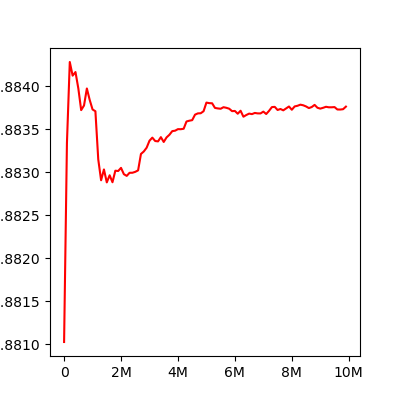}
    \includegraphics[scale=0.36]{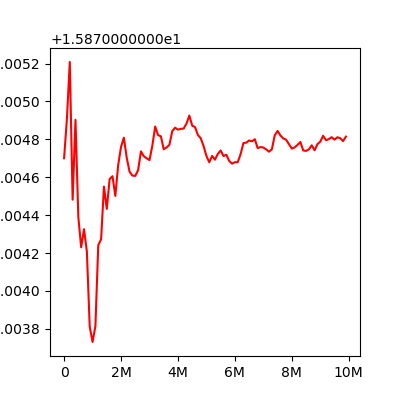}
    \\[\smallskipamount]
    \includegraphics[scale=0.36]{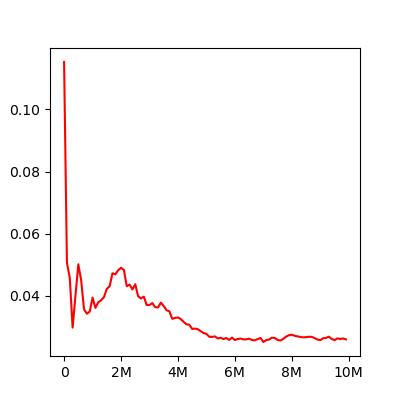}
    \includegraphics[scale=0.36]{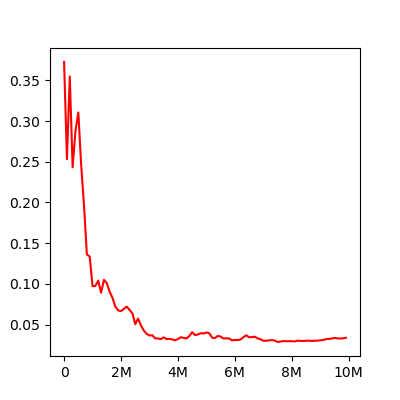}
    \includegraphics[scale=0.36]{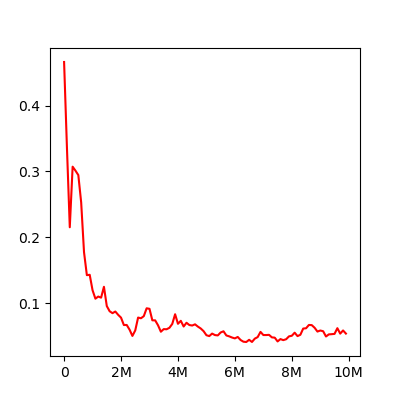}
    \includegraphics[scale=0.36]{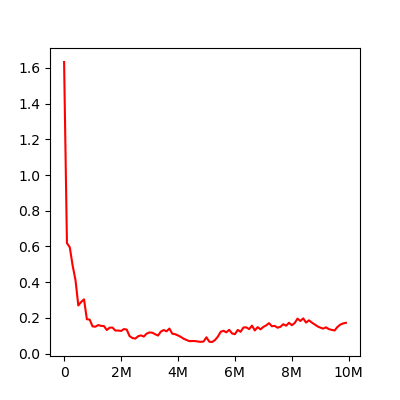}

\caption{Evolution of the Wasserstein Distance for eight different chains. Langevin Dynamics on the top row, Jump Langevin / Our method on the bottom row. The chains are estimating the posteriors with 2, 8, 16, 32 modes, respectively, on the columns.}
\label{fig:ddd}
\end{figure*}

Result show (Figure \ref{fig:ddd}) the inability of correctly expressing the posteriors when using pure Langevin Dynamics. Our method instead appears to effectively converge to the appropriate stationary distribution.

\section{Conclusion}
The two primary classes of methods for sampling, as based on solely zeroth-order, or first-order posterior information, exhibit particular comparative advantages and disadvantages that favor them for different problem classes. We have shown that by carefully combining the two mechanisms of gradient seeking together with proposal-defined jumps, we can sample from the challenging class of high dimensional and multimodal distributions. It can be considered that this resembles, e.g., trust region step acceptance rejection of gradient and Hessian generated iterate estimates in standard nonconvex optimization algorithms. As such, one could foresee a number of additional developments of modularly considering different gradient-based chains (i.e., HMC) and jump mechanisms (i.e., as including memory). This can open a broad research program of investing more challenging posterior distributions as the accuracy and robustness demands of contemporary AI systems advance.

\subsubsection*{References}
\vspace*{-25pt}
\bibliographystyle{abbrvnat}
\bibliography{biblio}

\begin{thebibliography}{18}
\providecommand{\natexlab}[1]{#1}
\providecommand{\url}[1]{\texttt{#1}}
\expandafter\ifx\csname urlstyle\endcsname\relax
  \providecommand{\doi}[1]{doi: #1}\else
  \providecommand{\doi}{doi: \begingroup \urlstyle{rm}\Url}\fi

\bibitem[Barbu and Zhu(2020)]{barbu2020monte}
A.~Barbu and S.~Zhu.
\newblock \emph{Monte Carlo Methods}.
\newblock Springer, 2020.
\newblock ISBN 9789811329708.

\bibitem[Chen et~al.(2020)Chen, Du, and Tong]{chen2020stationary}
X.~Chen, S.~S. Du, and X.~T. Tong.
\newblock On stationary-point hitting time and ergodicity of stochastic
  gradient langevin dynamics.
\newblock \emph{Journal of Machine Learning Research}, 2020.

\bibitem[Cheng and Bartlett(2018)]{ref3}
X.~Cheng and P.~Bartlett.
\newblock Convergence of langevin mcmc in kl-divergence.
\newblock In F.~Janoos, M.~Mohri, and K.~Sridharan, editors, \emph{Proceedings
  of Algorithmic Learning Theory}, volume~83 of \emph{Proceedings of Machine
  Learning Research}, pages 186--211. PMLR, 07--09 Apr 2018.

\bibitem[Dalalyan(2014)]{ref1}
A.~Dalalyan.
\newblock Theoretical guarantees for approximate sampling from smooth and
  log-concave densities.
\newblock \emph{Journal of the Royal Statistical Society: Series B (Statistical
  Methodology)}, 79, 12 2014.
\newblock \doi{10.1111/rssb.12183}.

\bibitem[Durmus and Moulines(2016)]{ref2}
A.~Durmus and E.~Moulines.
\newblock High-dimensional bayesian inference via the unadjusted langevin
  algorithm, 2016.

\bibitem[Green(1995)]{green1995reversible}
P.~J. Green.
\newblock Reversible jump markov chain monte carlo computation and bayesian
  model determination.
\newblock \emph{Biometrika}, 82\penalty0 (4):\penalty0 711--732, 1995.

\bibitem[Grenander and Miller(1994)]{grenander1994representations}
U.~Grenander and M.~I. Miller.
\newblock Representations of knowledge in complex systems.
\newblock \emph{Journal of the Royal Statistical Society: Series B
  (Methodological)}, 56\penalty0 (4):\penalty0 549--581, 1994.

\bibitem[Han et~al.(2004)Han, Tu, and Zhu]{han2004range}
F.~Han, Z.~Tu, and S.-C. Zhu.
\newblock Range image segmentation by an effective jump-diffusion method.
\newblock \emph{IEEE Transactions on pattern analysis and machine
  intelligence}, 26\penalty0 (9):\penalty0 1138--1153, 2004.

\bibitem[Izmailov et~al.(2021)Izmailov, Vikram, Hoffman, and Wilson]{bayespost}
P.~Izmailov, S.~Vikram, M.~D. Hoffman, and A.~G.~G. Wilson.
\newblock What are bayesian neural network posteriors really like?
\newblock In M.~Meila and T.~Zhang, editors, \emph{Proceedings of the 38th
  International Conference on Machine Learning}, volume 139 of
  \emph{Proceedings of Machine Learning Research}, pages 4629--4640. PMLR,
  18--24 Jul 2021.

\bibitem[Ma et~al.(2019)Ma, Chen, Jin, Flammarion, and Jordan]{samplopt}
Y.-A. Ma, Y.~Chen, C.~Jin, N.~Flammarion, and M.~I. Jordan.
\newblock Sampling can be faster than optimization.
\newblock \emph{Proceedings of the National Academy of Sciences}, 116\penalty0
  (42):\penalty0 20881--20885, 2019.
\newblock \doi{10.1073/pnas.1820003116}.

\bibitem[Mackay(1995)]{MacKay95}
D.~J.~C. Mackay.
\newblock Probable networks and plausible predictions — a review of practical
  bayesian methods for supervised neural networks.
\newblock \emph{Network: Computation in Neural Systems}, 6\penalty0
  (3):\penalty0 469--505, 1995.
\newblock \doi{10.1088/0954-898X\_6\_3\_011}.

\bibitem[Mengersen and Tweedie(1996)]{metroconvtweedie}
K.~L. Mengersen and R.~L. Tweedie.
\newblock {Rates of convergence of the Hastings and Metropolis algorithms}.
\newblock \emph{The Annals of Statistics}, 24\penalty0 (1):\penalty0 101 --
  121, 1996.
\newblock \doi{10.1214/aos/1033066201}.
\newblock URL \url{https://doi.org/10.1214/aos/1033066201}.

\bibitem[Neal(1996)]{nealbook}
R.~M. Neal.
\newblock \emph{Bayesian learning for neural networks}.
\newblock Springer Science \& Business Media, 1996.

\bibitem[Roberts and Tweedie(1996)]{roberts1996exponential}
G.~O. Roberts and R.~L. Tweedie.
\newblock Exponential convergence of langevin distributions and their discrete
  approximations.
\newblock \emph{Bernoulli}, pages 341--363, 1996.

\bibitem[Stroock and Varadhan(1997)]{stroock1997multidimensional}
D.~W. Stroock and S.~S. Varadhan.
\newblock \emph{Multidimensional diffusion processes}, volume 233.
\newblock Springer Science \& Business Media, 1997.

\bibitem[Wang(2020)]{metroconv2}
G.~Wang.
\newblock Exact convergence rate analysis of the independent
  metropolis-hastings algorithms, 2020.
\newblock URL \url{https://arxiv.org/abs/2008.02455}.

\bibitem[Welling and Teh(2011)]{wnt}
M.~Welling and Y.~W. Teh.
\newblock Bayesian learning via stochastic gradient langevin dynamics.
\newblock In \emph{Proceedings of the 28th International Conference on
  International Conference on Machine Learning}, ICML'11, page 681–688,
  Madison, WI, USA, 2011. Omnipress.
\newblock ISBN 9781450306195.

\bibitem[Yin and Zhu(2010)]{bookhybrid}
G.~Yin and C.~Zhu.
\newblock \emph{Hybrid switching diffusions. Properties and applications},
  volume~63.
\newblock 01 2010.
\newblock ISBN 978-1-4419-1106-3.
\newblock \doi{10.1007/978-1-4419-1105-6}.

\end{thebibliography}

\end{document}